\documentclass{article}

\usepackage{diagbox}
\usepackage{makecell}
\usepackage{booktabs}
\usepackage{tikz,tikz-3dplot}
\usepackage{bbm}
\usepackage{amsfonts}
\usepackage{amsmath}
\usepackage{amsthm}
\usepackage{url,ifthen}
\usepackage[shortlabels]{enumitem}
\usepackage{srcltx}
\usepackage{dsfont}
\usepackage{multirow}
\usepackage{boxedminipage}
\usepackage[margin=1.15in]{geometry}
\usepackage{nicefrac}
\usepackage{xspace}
\usepackage{graphicx}
\usepackage{color}
\usepackage{subfigure}
\usepackage{colortbl}
\usepackage{setspace}
\usepackage{pgfplots}
\pgfplotsset{compat=1.12}
\usepackage{algorithm}
\usepackage[noend]{algorithmic}

\usepackage{xcolor}
\definecolor{DarkGreen}{rgb}{0.1,0.5,0.1}
\definecolor{DarkRed}{rgb}{0.5,0.1,0.1}
\definecolor{DarkBlue}{rgb}{0.1,0.1,0.5}
\definecolor{Gray}{rgb}{0.2,0.2,0.2}

\definecolor{c1}{RGB}{38, 70, 83}
\definecolor{c2}{RGB}{42, 157, 143}
\definecolor{c3}{RGB}{233, 196, 106}
\definecolor{c5}{RGB}{231, 111, 81}
\definecolor{c4}{RGB}{244, 162, 97}

\definecolor{base}{rgb}{0.23299120924703914, 0.639586552066035, 0.9260706093977744}
\definecolor{instruct}{rgb}{0.21044753832183283, 0.6773105080456748, 0.6433941168468681}
\definecolor{instruct}{RGB}{229, 128, 145}
\definecolor{census}{rgb}{0.3126890019504329, 0.6928754610296064, 0.1923704830330379}
\definecolor{uniform}{rgb}{0.9333333333333333, 0.5215686274509804, 0.2901960784313726}

\usepackage{listings}
\lstdefinestyle{mystyle}{
    commentstyle=\color{DarkBlue},
    keywordstyle=\color{DarkRed},
    numberstyle=\tiny\color{Gray},
    stringstyle=\color{DarkGreen},
    basicstyle=\footnotesize,
    breakatwhitespace=false,         
    breaklines=true,                 
    captionpos=b,                    
    keepspaces=true,                 
    numbers=left,                    
    numbersep=5pt,                  
    showspaces=false,                
    showstringspaces=false,
    showtabs=false,                  
    tabsize=2
}
\usepackage{caption}

\usepackage[pdftex]{hyperref}
\hypersetup{
    unicode=false,          
    pdftoolbar=true,        
    pdfmenubar=true,        
    pdffitwindow=false,      
    pdfnewwindow=true,      
    colorlinks=true,       
    linkcolor=DarkBlue,          
    citecolor=DarkGreen,        
    filecolor=DarkRed,      
    urlcolor=DarkBlue,          
    pdftitle={},
    pdfauthor={},
}

\def\draft{1}

\def\submit{0}

\ifnum\draft=1 
    
\else
    
\fi

\ifnum\submit=1
\newcommand{\forsubmit}[1]{#1}
\newcommand{\forreals}[1]{}
\else
\newcommand{\forreals}[1]{#1}
\newcommand{\forsubmit}[1]{}
\fi

\newtheorem{theorem}{Theorem}[section]

\newtheorem{definition}[theorem]{Definition}

\newtheorem*{definition*}{Definition}

\theoremstyle{definition}

\newtheoremstyle{example_contd}
{\topsep} {\topsep}%
{}
{}
{\bfseries}
{.}
{1em}
{\thmname{#1} \thmnumber{ #2}\thmnote{#3} (continued)}

\theoremstyle{example_contd}

\newcommand{\chapterref}[1]{\hyperref[ch:#1]{Chapter~\ref{ch:#1}}}

\newcommand{\claimref}[1]{\hyperref[claim:#1]{Claim~\ref{claim:#1}}}

\newcommand{\corollaryref}[1]{\hyperref[cor:#1]{Corollary~\ref{cor:#1}}}

\newcommand{\definitionref}[1]{\hyperref[def:#1]{Definition~\ref{def:#1}}}

\newcommand{\equationref}[1]{\hyperref[eq:#1]{Equation~\ref{eq:#1}}}

\newcommand{\factref}[1]{\hyperref[fact:#1]{Fact~\ref{fact:#1}}}

\newcommand{\figureref}[1]{\hyperref[fig:#1]{Figure~\ref{fig:#1}}}

\newcommand{\tableref}[1]{\hyperref[tab:#1]{Table~\ref{tab:#1}}}

\newcommand{\itemref}[1]{\hyperref[item:#1]{Item~(\ref{item:#1})}}

\newcommand{\lemmaref}[1]{\hyperref[lem:#1]{Lemma~\ref{lem:#1}}}

\newcommand{\propref}[1]{\hyperref[prop:#1]{Proposition~\ref{prop:#1}}}

\newcommand{\propositionref}[1]{\hyperref[prop:#1]{Proposition~\ref{prop:#1}}}

\newcommand{\remarkref}[1]{\hyperref[rem:#1]{Remark~\ref{rem:#1}}}

\newcommand{\sectionref}[1]{\hyperref[sec:#1]{Section~\ref{sec:#1}}}

\newcommand{\theoremref}[1]{\hyperref[thm:#1]{Theorem~\ref{thm:#1}}}

\usepackage[charter]{mathdesign}
\makeatletter
\DeclareFontFamily{OMS}{cmsy}{\skewchar\font48}
\DeclareFontShape{OMS}{cmsy}{m}{n}{
  <5><6><7><8><9><10><10.95><12><14.4><17.28><20.74><24.88>gen*cmsy
}{}
\DeclareMathAlphabet{\mathcmcal}{OMS}{cmsy}{m}{n}
\makeatother

\usepackage{microtype}

\renewcommand{\hat}{\widehat}

\renewcommand{\leq}{\leqslant}

\renewcommand{\geq}{\geqslant}

\usepackage{bm}

\newcommand{\remove}[1]{}

\usepackage{hyperref}
\usepackage{graphicx}
\usepackage{caption}
\usepackage{xcolor}
\usepackage{xspace}
\usepackage{xurl}
\usepackage{tabularx} 
\usepackage[utf8]{inputenc} 

\usepackage[T1]{fontenc}    
\usepackage{hyperref}       
\usepackage{url}            
\usepackage{booktabs}       
\usepackage{nicefrac}       
\usepackage{microtype}      
\usepackage{xcolor}         

\definecolor{myblue}{rgb}{0.23299120924703914, 0.639586552066035, 0.9260706093977744}
\definecolor{myred}{rgb}{0.9677975592919913, 0.44127456009157356, 0.5358103155058701}
\definecolor{mygreen}{rgb}{0.3126890019504329, 0.6928754610296064, 0.1923704830330379}
\definecolor{myorange}{rgb}{0.9333333333333333, 0.5215686274509804, 0.2901960784313726}
\definecolor{myorange2}{rgb}{0.8352941176470589, 0.3686274509803922, 0.0}

\usepackage{tikz}
\usetikzlibrary{positioning,arrows.meta}

\usepackage{fontawesome}
\usepackage[round]{natbib}
\usepackage{soul}
\usepackage{pifont}
\usepackage[symbol]{footmisc}
\usepackage{authblk}
\usepackage{url}
\usepackage{longtable}
\title{Computational Arbitrage in AI Model Markets}

\author[]{Ricardo Olmedo}
\author[]{Bernhard Sch\"olkopf}
\author[]{Moritz Hardt}
\affil[]{Max Planck Institute for Intelligent Systems, T\"ubingen}

\usepackage{longtable}
\usepackage{graphicx}

\usepackage{array}
\usepackage{multirow}

\usepackage{pifont}
\usepackage{threeparttable}

\newcounter{daggerfootnote}

\begin{document}

\maketitle
\setcounter{footnote}{0}
\renewcommand{\thefootnote}{\arabic{footnote}}

\begin{abstract}
  Consider a market of competing model providers selling query access to models with varying costs and capabilities. Customers submit problem instances and are willing to pay up to a budget for a verifiable solution. An arbitrageur efficiently allocates inference budget across providers to undercut the market, thus creating a competitive offering with no model-development risk. In this work, we initiate the study of arbitrage in AI model markets, empirically demonstrating the viability of arbitrage and illustrating its economic consequences. We conduct an in-depth case study of SWE-bench GitHub issue resolution using two representative models, GPT-5 mini and DeepSeek v3.2. In this verifiable domain, simple arbitrage strategies generate net profit margins of up to 40\%. Robust arbitrage strategies that generalize across different domains remain profitable. Distillation further creates strong arbitrage opportunities, potentially at the expense of the teacher model’s revenue. Multiple competing arbitrageurs drive down consumer prices, reducing the marginal revenue of model providers. At the same time, arbitrage reduces market segmentation and facilitates market entry for smaller model providers by enabling earlier revenue capture. Our results suggest that arbitrage can be a powerful force in AI model markets with implications for model development, distillation, and deployment\footnote{Code, data, and models available at https://github.com/RicardoDominguez/computational-arbitrage}.
\end{abstract}

\section{Introduction}

Industry experts call it a “Cambrian explosion”---the rapid growth of competing AI models in the marketplace \citep{chernova2025openrouter}. As of January 2026, Open Router provides API access to more than 600 different models, while Microsoft Foundry lists more than 11000 models. Customers face myriad model choices of varying costs and capabilities from all kinds of providers. From the customer’s perspective, massive API markets are therefore particularly appealing for workflows with verifiable solutions, such as software passing a test suite. When solutions are verifiable, customers can choose freely between different options without the need for familiarity or trust in the model provider.

At the same time, the complex fragmentation of the AI API market creates potential for \emph{arbitrage}. The same level of performance implicitly trades at different prices across different API endpoints. Solving half of all SWE-bench problems costs roughly \$10 with GPT-5 mini and \$20 with DeepSeek v3.2. However, DeepSeek typically solves harder problems in fewer attempts than GPT-5. At a higher target of a 75\% SWE-bench solve rate, DeepSeek is the cheaper model, requiring \$120 compared to over \$150 for GPT-5 mini.

In this work, we study how an arbitrageur may exploit such cost asymmetries for profit. A simple arbitrage strategy first queries GPT-5 mini some number of times before switching over to DeepSeek. Easier problems are thus cheaply solved by GPT-5 mini, while more difficult ones are ultimately handled by DeepSeek. By choosing a good switch-over point, the arbitrageur can purchase a target level of performance at a lower cost than either model alone. This cost advantage is in turn a profit opportunity (see Figure 1).  The strategy requires no up-front investment and is formally risk-free when customers commit to a no-refund budget. At worst, the arbitrageur fails to create a competitive product and ends up with zero profit, but no losses.

Arbitrage is fundamental to financial markets, yet we lack a corresponding understanding of its computational counterpart in the context of AI model markets. Initiating the study of computational arbitrage in AI model markets, we empirically demonstrate the viability of arbitrage in a realistic setting and illustrate its potential economic consequences. Our results suggest that arbitrage in AI model markets may have powerful implications for model development, distillation, and deployment. 

In more detail, our contributions are:

\begin{figure*}[t]
    \centering
    \includegraphics[width=\linewidth]{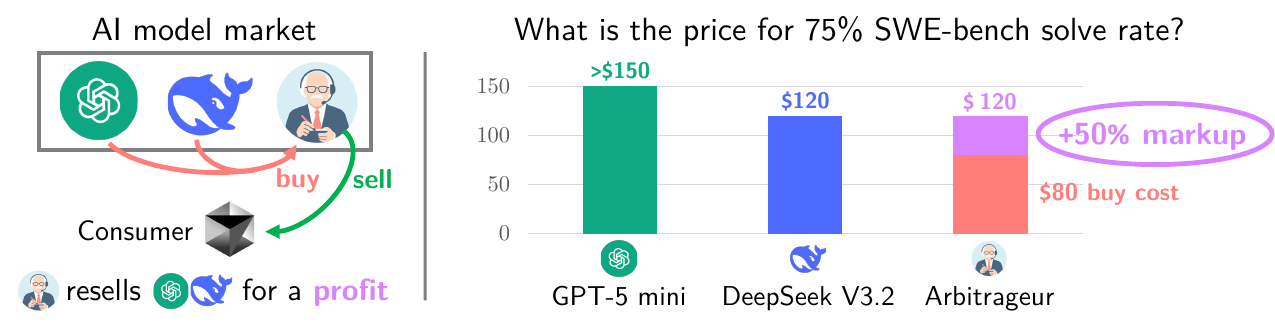}
    \caption{Consider a model market with three providers: GPT-5 mini, DeepSeek v3.2, and an arbitrageur. Consumers demand a target level of performance on SWE-bench–type tasks; for example, a 75\% SWE-bench solve rate. Through repeated sampling, GPT-5 mini and DeepSeek each achieve a 75\% SWE-bench solve rate at costs of \$150 and \$120, respectively. The arbitrageur instead sources generations by first querying GPT-5 mini (at up to \$0.08 per problem) and, if that fails, querying DeepSeek. Using this strategy, the arbitrageur attains the same 75\% solve rate at a cost of \$80. This cost advantage creates a profit opportunity: the arbitrageur can resell its sourced generations at markups of up to 50\% while still undercutting the market.}
    \label{fig:figintro}
\end{figure*}

\begin{itemize}
    \item \textbf{We formalize the concept of computational arbitrage in AI model markets.} Arbitrage opportunities arise when a market participant can simultaneously buy and sell API calls from a combination of multiple providers at a profit, while incurring no model-development risk.
    \item \textbf{We demonstrate the feasibility of computational arbitrage} via an in-depth case study of SWE-bench GitHub issue resolution with GPT-5 mini and DeepSeek v3.2 models. In this setting, simple arbitrage strategies yield net profit margins of up to 40\%. We additionally demonstrate that arbitrage is robust: arbitrage policies are inexpensive to fit and remain profitable under distribution shifts.
    \item \textbf{We analyze the economic implications of computational arbitrage.} Competition among arbitrageurs drives down consumer prices, at the expense of providers’ marginal revenues. At the same time, arbitrage reduces market segmentation, even allowing smaller models to capture some of the revenue generated by consumer demand for frontier model performance.
    \item \textbf{Model distillation gives rise to arbitrage opportunities}. Through our own scaling experiments, we show that increased distillation consistently improves cost-to-solution, thereby creating increasingly profitable arbitrage opportunities. We then show that distillation can directly undermine the teacher model’s revenue, potentially eliminating it altogether. To do so, we train mini-coder 4B, a small model that outperforms Qwen Coder 30B in terms of cost-to-solution.
\end{itemize}

\section{Computational arbitrage}

We consider an AI model marketplace in which consumers can choose among multiple model providers. Providers take in queries $x \in \mathcmcal{X}$ (e.g., a software issue description) together with an inference budget $b\in\mathbb{R}_{+}$. They then return some output $y \in \mathcmcal{Y}$ (e.g., a proposed fix) while charging the consumer some cost~$c \leq b$. Formally, we model each provider $p$ as a conditional distribution $p(y, c \mid x, b)$. Consumers derive utility from providers’ outputs and will seek to query the provider that offers the best trade-off between utility and cost.

We evaluate market providers using a standardized performance metric  $u: \mathcmcal{X} \times\mathcmcal{Y} \rightarrow \mathbb{R}$ (e.g., accuracy), and assume that consumer utility increases monotonically with model performance. For a given query distribution $x\sim D$ and provider~$p$, varying the inference budget $b$ induces different trade-offs between expected cost $\bar{c}_p$ and expected performance $\bar{u}_p$, or \emph{cost} and \emph{performance} for short, specifically
\begin{equation}
    \bar{c}_p(b) = \mathbb{E}_{x \sim D,\; c\sim p(\cdot \mid x,b)}\left[c\right] \text{,}\quad \bar{u}_p(b) = \mathbb{E}_{x \sim D,\; y\sim p(\cdot \mid x,b)}\left[u(x, y)\right].
\end{equation}

The minimum cost required for a provider $p$ to achieve a target performance level $u$ is
\begin{equation}
    C_p(u) =\min_{b \;\text{s.t.}\; \bar{u}_p(b) \geq u} \bar{c}_p(b).
\end{equation}
A fully informed, rational consumer will choose the lowest-cost provider. For a market of providers \mbox{$\mathbf{P}:=\left\{p_1, p_2, \ldots\right\}$}, we define the \emph{market price} $C_{\mathbf{P}}(u)$ for performance level $u$ as the minimum cost at which any provider offers that level of performance:  
\begin{equation}
C_{\mathbf{P}}(u) = \min_{p \in \mathbf{P}} C_p(u).
\end{equation}

As we will see, arbitrageurs seek to obtain below-market prices, thereby creating profit opportunities.

\paragraph{Computational arbitrage.} An arbitrageur is a market participant who resells other providers' outputs for a profit. Specifically, given a query $x\in\mathcmcal{X}$ and a budget~$b\in\mathbb{R}$, an arbitrageur purchases one or more model responses from the market $\mathbf{P}$, incurring some cost~$c\in\mathbb{R}_{+}$. The arbitrageur then returns one of the acquired responses $y\in\mathcmcal{Y}$ to the consumer, applying some cost markup~$\delta > 0$. We abstract the arbitrageur's policy for sourcing generations from the market as a conditional distribution~$q(y, c \mid x, b)$.

An arbitrageur cannot operate at a loss. At worst, it is unable to offer competitive prices, thus failing to attract demand and earning zero profit. That is, computational arbitrage is risk-free by construction. To profit, however, arbitrageurs must achieve prices lower than those otherwise available in the market. 

\begin{definition}[Arbitrage Opportunity]
An \emph{arbitrage opportunity} exists in a marketplace $\mathbf{P}$ under a query distribution $D$ if there exists an arbitrage policy $q$ such that the policy achieves some level of performance at a cost strictly lower than its market price. Formally,
\begin{equation}
\exists\;\text{arbitrage policy } q, u\in\mathbb{R} \quad \text{s.t.} \quad 
C_q(u) < C_{\mathbf{P}}(u),
\label{eq:arbitrage}
\end{equation}
where $C_q(u)$ denotes the arbitrageur’s expected cost of achieving the target performance level $u$, and $C_{\mathbf{P}}(u)$ denotes the market price for the same performance level.
\end{definition}

Intuitively, the arbitrageur can capture the spread between the prevailing market price $C_{\mathbf{P}}(u)$ for the level of performance $u$ and its cost $C_{q}(u)$ of purchasing that same level of performance. Specifically, for any given perfor mance level $u$, an arbitrage policy $q$ can earn a marginal profit of 
\begin{equation}
    \Pi_q(u) = \max\left(C_{\mathbf{P}}\left(u\right)-C_q(u), 0\right) \geq 0.
\label{eq:profit}
\end{equation}
By construction, arbitrageurs cannot operate at a loss, since at worst they simply fail to attract any demand. 

Arbitrageurs seek to maximize profit. The arbitrage policy~$q^*$ that maximizes profit in the market is
\begin{equation}
    q^* =  \underset{q}{\arg\max} \;\;\int_{\mathbb{R}} \Pi_q(u) w(u)\;du,
    \label{eq:obj}
\end{equation}
where the weighting function $w(u)$ captures the market demand for any given level of performance~$u$. For simplicity, we will assume that consumer demand is uniform across all performance levels, that is, $w(u) = 1$.

In summary, arbitrageurs achieve below-market prices by optimally sourcing generations from the market, thereby creating opportunities for profit. In the next section, we present an empirical study of computational arbitrage in software issue resolution.

\section{Arbitrage in software issue resolution}

We focus on SWE-bench Verified~\citep{swebench, verified}, the leading benchmark for software issue resolution. It comprises 500 software issues sourced from GitHub, each paired with unit tests to verify the functional correctness of model-generated patches. Model performance is measured as the fraction of issues for which the model produces a successful patch. 
For our initial exposition, we compare GPT-5 mini~\citep{singh2025openai} and DeepSeek v3.2 Thinking~\citep{liu2025deepseek}. We conduct additional experiments for Lean4 formal threorem proving, see Appendix~\ref{app:lean}.


We scale the inference budget through repeated sampling. Specifically, a model is repeatedly queried to solve a target software issue until it produces a successful patch or exhausts its inference budget~\citep{humaneval, li2022competition}. Further details on the evaluation set-up and model pricing are in Appendices~\ref{app:eval} and~\ref{app:price}.

For each model $i$ and issue $j$, we observe $n_{ij}$ solution attempts and $m_{ij}$ correct solutions, along with the mean cost per attempt $\hat{s}_{ij}$.
From these, we estimate the probability that the issue is solved within $k$ independent attempts using the standard unbiased estimator $\mathrm{pass}@k = 1 - \binom{n-m}{k} / \binom{n}{k}$.
To express performance in terms of monetary cost rather than number of attempts, we convert a dollar budget~$b$ into an equivalent number of attempts $k = b / \hat{s}_{ij}$, yielding a per-issue performance curve $u_{ij}(b) = \mathrm{pass}@(b/\hat{s}_{ij})$. Aggregating across issues gives the model's expected solve rate at budget~$b$, that is, $\bar{u}_i(b) = \frac{1}{|J|} \sum_{j \in J} u_{ij}(b)$.

\begin{figure*}[t]
    \centering
    \includegraphics[width=\linewidth]{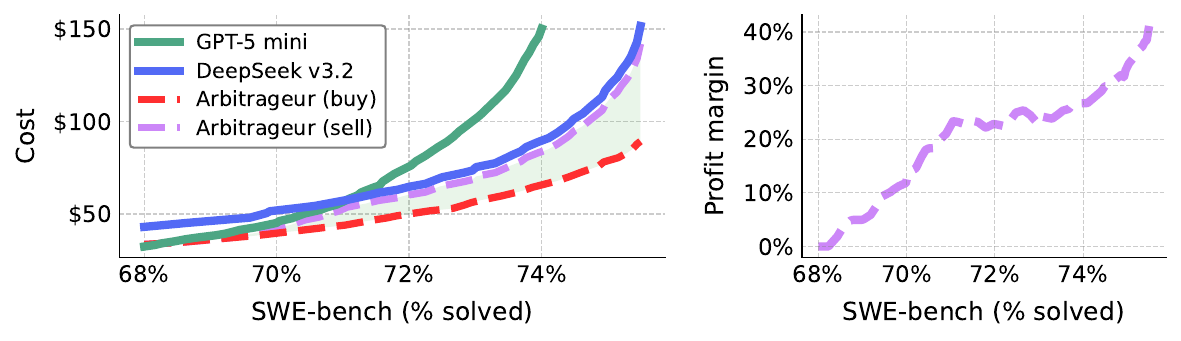}
    \caption{Inference cost for GPT-5 mini and DeepSeek v3.2 to reach different SWE-bench performance levels, with inference budgets scaled through repeated sampling (up to \$1 per issue). We also evaluate the following arbitrage policy: allocate up to \$0.08 to GPT-5 mini and, if it fails, spend the remaining \$0.92 on DeepSeek. The arbitrage policy (red) achieves solve rates above 68\% at a lower cost than either GPT-5 mini or DeepSeek. This cost advantage enables the arbitrageur to profit by reselling its generations close to market price (purple).}
    \label{fig:fig1}
\end{figure*}

We then compute each model’s expected cost at different budgets $b$. Specifically, the expected total cost is $c_i(b) = |J| \int_0^b \bigl(1 - \bar{u}_i(x)\bigr)\, dx$, which follows from the survival-function identity for non-negative random variables. We plot in Figure~\ref{fig:fig1} (left) inference cost $c_i$ versus performance $\bar{u}_i$ for {GPT-5} mini and DeepSeek. {GPT-5} mini is more cost-efficient for lower budgets, while DeepSeek is preferable for higher budgets.

\paragraph{Arbitrage policy.}

We construct arbitrage policies using a model cascade design~\citep{varshney2022model}, in which market providers are queried sequentially in a fixed order until a successful patch is obtained. The central challenge is determining how to allocate the inference budget optimally across the model cascade.

We denote by~$\tau_i \in \mathbb{R}$ the cap on spending for provider~$p_i$. Given a total budget~$b$, provider~$i$ is allocated
\begin{equation}
    b_i^{(\tau)} = \min\Bigl(\max\bigl(b - \textstyle\sum_{k<i}\tau_k,\; 0\bigr),\; \tau_i\Bigr),
\end{equation}
i.e., whatever remains of~$b$ after the preceding providers have each claimed up to their cap, clamped to~$[0, \tau_i]$.

Each provider independently attempts every unsolved issue using its allocated budget, so the probability that issue~$j$ is solved by at least one provider in the cascade is
\begin{equation}
    u_j^{(\tau)}(b) \;=\; 1 - \prod_{i=1}^{|I|}\bigl(1 - u_{i,j}(b_i^{(\tau)})\bigr),
\end{equation}
where $u_{i,j}(b_i)$ is the solve probability of model~$i$ on issue~$j$ when given budget~$b_i$. Averaging over issues yields the cascade's expected performance at budget~$b$, that is, 
    $\bar{u}^{(\tau)}(b) \;=\; \frac{1}{|J|}\sum_{j \in J} u_j^{(\tau)}(b)$.

As shown earlier, GPT-5 is more cost-effective than DeepSeek for low budgets. Therefore, we build the cascade by first querying GPT-5 mini and then DeepSeek. Specifically, we search for the profit-maximizing allocation $\tau^*$ according to Equation~\ref{eq:obj}. Under a maximum inference budget of \$1, this yields $\tau^* = \left\{\$0.08, \$0.92\right\}$, meaning that the arbitrageur allocates up to \$0.08 per issue to GPT-5 mini before querying DeepSeek.

We compare the cost-performance of the arbitrageur against GPT-5 mini and DeepSeek in Figure~\ref{fig:fig1} left. The arbitrageur (red curve) can source any given level of performance above 68\% solve rate at a cheaper cost than either of the two models. These efficiency gains generate opportunities for profit, as the arbitrageur can undercut the market by pricing its outputs slightly below the market price (pink line). By following this strategy, the arbitrageur can achieve profit margins of up to 40\%, as shown in Figure~\ref{fig:fig1} right. The remarkable profitability of arbitrage highlights the inefficiency of querying either GPT-5 mini or DeepSeek alone.

\begin{figure*}[t]
    \centering
    \includegraphics[width=\linewidth]{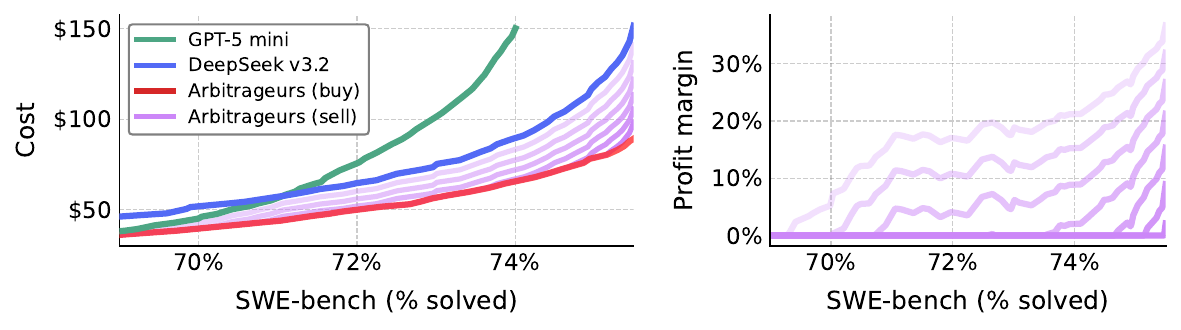}
    \caption{Two arbitrageurs deploy the same arbitrage policy but compete on price. They take turns updating their prices to undercut each other. Earlier turns are plotted with greater transparency. \emph{Left:} Competition between arbitrageurs drives market prices downward. In equilibrium, the market price equals the arbitrageurs' buy price. \emph{Right:} While arbitrage is initially highly profitable, profit opportunities eventually vanish.}
    \label{fig:competition}
\end{figure*}

\subsection{Economic implications of computational arbitrage}

\subsubsection{Competition between arbitrageurs reduces prices}

We have shown how an arbitrageur can profit by entering an inefficient market. However, even after this entry, arbitrage opportunities may still remain. In particular, a competing arbitrageur could enter the market and undercut the first by accepting a smaller profit margin.

Consider two arbitrageurs who source their outputs from the profit-maximizing arbitrage policy~$q^*$, and compete over pricing. The two arbitrageurs take turns updating their prices, so as to be cheaper than the prevailing market price. We plot in Figure~\ref{fig:competition} left the market's cost-performance frontier as the two arbitrageurs sequentially update their prices. By undercutting each other, market prices reduce considerably. As a result, arbitrage profitability decreases, as plotted in Figure~\ref{fig:competition} right, and arbitrage ultimately ceases to be profitable.

These dynamics correspond to the classic framework of Bertrand competition~\citep{mas1995microeconomic}. When two providers have identical marginal costs, offer identical products, and consumers have perfect information of providers' prices, the equilibrium outcome is one in which market price equals marginal cost. By construction, the two competing arbitrageurs share identical marginal costs~$C_{q^*}$. Consequently, the equilibrium market price of the new market $\mathbf{P}'$ is at most the arbitrageurs' marginal cost, that is,
\begin{equation}
C_{\mathbf{P}'}(u) = \min\left(C_{\mathbf{P}}(u), C_{q^*}(u)\right).
\end{equation}

As a result, the profit-maximizing arbitrage policy $q^*$ for the original market $\mathbf{P}$ ceases to be profitable in the new market $\mathbf{P}'$. In this sense, arbitrage is self-defeating. When arbitrage opportunities exist, competition among arbitrageurs quickly eliminates them. From the consumer side, the consequence is lower market prices. We next examine the economic implications of arbitrage for model providers.

\begin{figure*}[t]
    \centering
    \includegraphics[width=\linewidth]{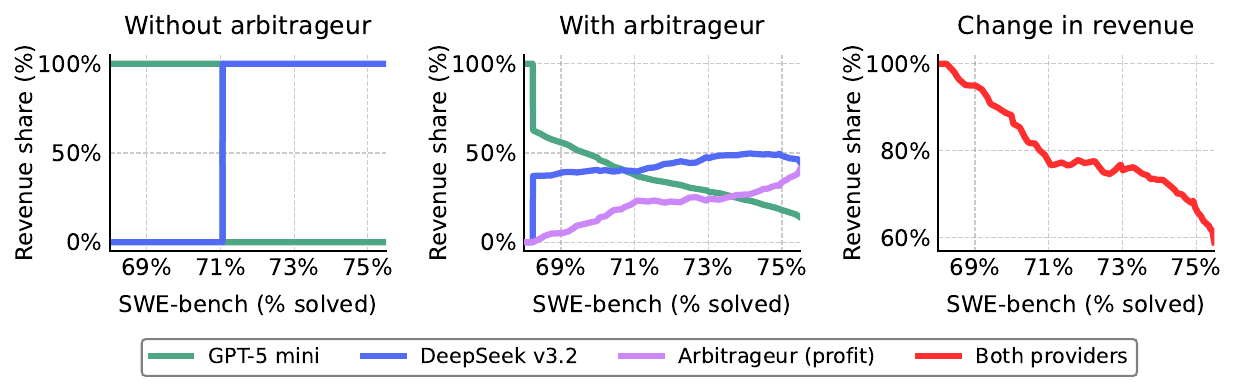}
    \caption{Revenue split across model providers for different levels of performance. \emph{Left:} In the absence of arbitrageurs, the market is segmented by performance, with a single model dominating each segment. \emph{Middle:} Arbitrageurs eliminate this segmentation, allowing both models to earn revenue across a much broader range of levels of performance. \emph{Right:} Arbitrageurs reduce providers’ marginal revenue, with the lost surplus transferred to arbitrageur profits or passed on to consumers as lower market prices.}
    \label{fig:revenuesplit}
\end{figure*}

\subsubsection{Arbitrage breaks market segmentation and reduces providers' revenue}

In this section, we study the implications of computational arbitrage for market segmentation. As discussed previously, GPT-5 mini is more cost-effective for lower inference budgets, whereas DeepSeek is more cost-effective for higher budgets. This implies that the market is segmented into two distinct performance tiers, as shown in Figure~\ref{fig:revenuesplit} left.  Consumers seeking performance below a 71\% SWE-bench solve rate should query GPT-5 mini, while those seeking higher performance should query DeepSeek.

When the arbitrageur enters the market, it captures all consumer demand by offering lower prices. Nevertheless, both providers may continue to earn revenue, since the arbitrageur relies on them for sourcing its outputs. We plot in Figure~\ref{fig:revenuesplit} middle how consumer expenditure is split across providers once the arbitrageur enters the market. Notably, market segmentation disappears. Instead, DeepSeek earns revenue across a wider range of SWE-bench performance levels. Similarly, GPT-5 mini earns revenue across the entire performance spectrum, including at the frontier (i.e., a 75\% SWE-bench solve rate).

This latter observation has important implications. In markets where consumers only seek frontier performance, cheaper models such as GPT-5 mini are not irrelevant. On the contrary, by contributing to overall efficiency, cheap models can earn revenue even at the performance frontier. Therefore, successful market entry does not require offering the best-performing model; being sufficiently cheap can suffice.

Arbitrage profits come at the expense of providers' marginal revenue. In our setting, marginal revenue decreases by up to 40\%, as plotted in Figure~\ref{fig:revenuesplit} right. This revenue loss is either transformed into arbitrage profits or, in the presence of competing arbitrageurs, passed on to consumers as lower prices\footnote{Note that an overall reduction in prices may result in higher trading volumes, and thus larger total revenue.}.

\subsection{Arbitrage is inexpensive and robust}
As a theoretical concept in economics, arbitrage should require no initial investment and entail no risk~\citep{hulloptions}. In practice, however, arbitrageurs necessarily incur costs~\citep{limitsarbitrage}. We now show that computational arbitrage is practical: profitable policies are inexpensive to find and generalize well.

\begin{figure*}[t]
    \centering
    \includegraphics[width=\linewidth]{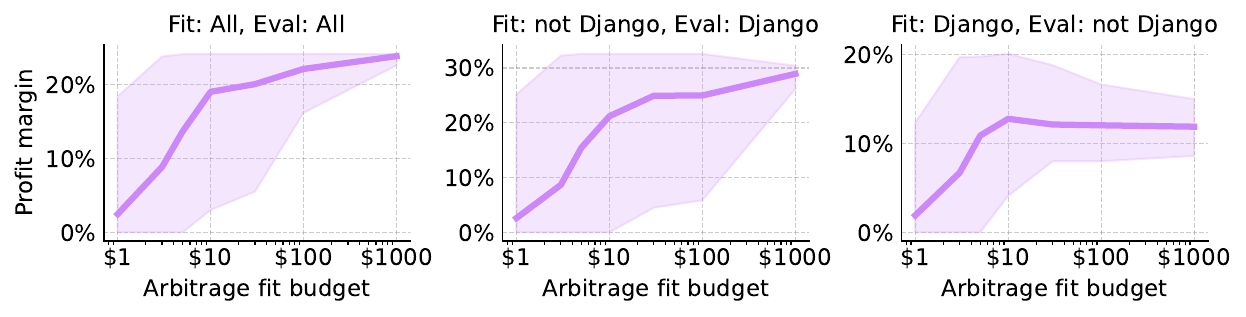}
    \caption{Profit margin across different search budgets when fitting the arbitrage policy. The solid line represents mean profitability, whereas the shaded area indicates the 95\% confidence interval, computed by bootstrapping over the samples acquired within the search budget. \emph{Left:} When fitting a fixed query distribution, small search budgets (e.g., \$10) consistently yield profitable arbitrage policies. \emph{Middle and right:} We fit the arbitrageur either on software issues from the Django library or on issues from other repositories, and evaluate the resulting arbitrage policy on the held-out data. We find that, on expectation, the arbitrageur remains profitable under such distribution shifts in the query distribution.}
    \label{fig:pricerob}
\end{figure*}

\paragraph{Cost of search.} Identifying arbitrage opportunities requires collecting a dataset of cost comparisons across a number of input queries. This dataset is then used to fit an arbitrage policy that maximizes expected profit. The cost comparisons can be collected while serving GPT-5 mini, with the only additional search cost arising from redundant queries to DeepSeek. We allow a search budget of \$0.5 per query. Consequently, in the worst case, a total search budget of \$10 permits only \$10 / \$0.50 = 20 price comparisons.

We plot in Figure~\ref{fig:pricerob} (left) the profitability of the fitted policy as a function of the search budget. We report mean profitability over the 70\% to 75\% SWE-bench performance range. The solid line represents mean profitability, whereas the shaded area indicates the 95\% confidence interval, computed by bootstrapping over the samples acquired within the search budget. In expectation, budgets as low as \$1 suffice to yield profitable arbitrage policies. However, the 95\% confidence intervals are wide due to the small sample sizes. A slightly larger search budget of \$10 allows for consistently fitting a profitable policy. Therefore, the initial investment required for computational arbitrage is minimal.

\paragraph{Robustness to the query distribution.} We split SWE-bench into issues from the Django repository and issues from all other repositories. We select Django because it accounts for roughly half of all SWE-bench issues. This split induces a natural distribution shift in the query distribution: Django primarily concerns web development, which differs substantially from other SWE-bench domains (e.g., scientific computing with scikit-learn). It also introduces a difficulty shift, as Django issues tend to be easier for LLMs to solve.

We fit arbitrage policies on one of the SWE-bench splits, and evaluate their profitability on the other split. We plot in Figure~\ref{fig:pricerob} (middle and right) out-of-distribution profitability against in-distribution search cost. We report mean profitability on the upper end of model performance, that is, 75\%-80\% solve rate for Django issues, and 61\%-66\% solve rate for non-Django issues. On expectation, arbitrage policies remain profitable even with search expenditures as low as \$1. At larger search budgets (e.g., \$30), the learned policies are consistently profitable. That is, the profit-maximizing arbitrage policy for each query distribution generalizes and remains profitable under reasonably large distribution shifts.

\subsection{Arbitrage in larger model markets}

So far, we have demonstrated arbitrage opportunities in a two-provider market. We now examine how these opportunities change when four additional model providers of varying sizes enter the market: Qwen 3 Coder~30B and 480B~\citep{qwencoder, yang2025qwen3}, Claude Sonnet 4.5~\citep{sonnet}, and our distilled mini-coder~4B model, trained using the distillation procedure described in Section~\ref{sec:distill}.

\begin{figure*}[t]
    \centering
    \includegraphics[width=\linewidth]{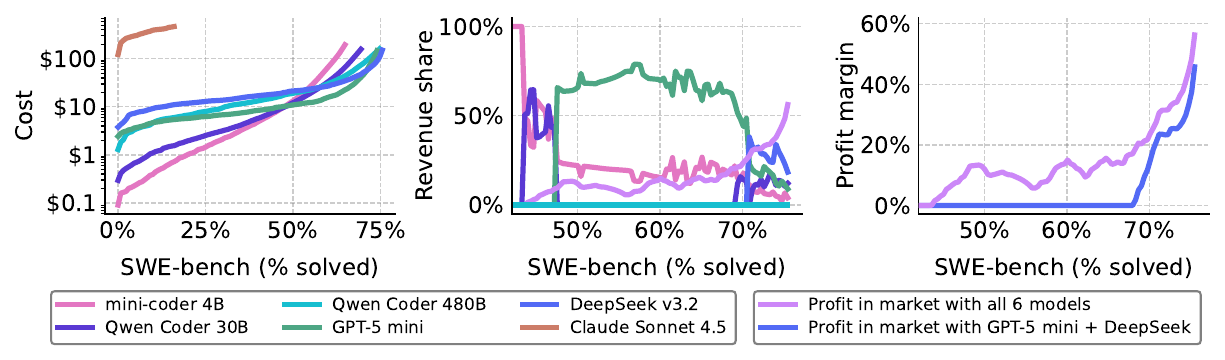}
    \caption{Model market with six providers of varying sizes. \emph{Left:} Cost for each model to achieve different levels of SWE-bench performance. \emph{Middle:} Revenue split across providers. There is little market segmentation, with up to four providers sharing revenue at a given performance level. Some models, such as Qwen Coder 480B and Claude Sonnet 4.5, are not competitive. \emph{Right:} Compared with the earlier two-model market (GPT-5 mini and DeepSeek), the six-model market yields more, and more profitable, arbitrage opportunities.}
    \label{fig:moremodels}
\end{figure*}

We plot the cost–performance curves of the six models in Figure~\ref{fig:moremodels} (left). Compared to GPT-5 mini and DeepSeek, mini-coder and Qwen Coder 30B are considerably smaller and therefore more efficient at low inference budgets. In contrast, Claude Sonnet 4.5 is too expensive in the compute regimes we consider; a \$1 budget is often insufficient to submit a solution for many SWE-bench problems. Finally, Qwen Coder 480B is dominated by GPT-5 mini and DeepSeek, as it is neither particularly cost-efficient nor high-performing.

We next examine the revenue earned by each model in the market. For each performance level between 45\% and 75\% SWE-bench solve rate, we search for the arbitrage policy with the lowest purchase cost. This cost represents the arbitrage-free (i.e., equilibrium) market price for that level of performance. We plot the corresponding revenue shares in Figure~\ref{fig:moremodels} (middle). The arbitrage-free market is not segmented: four of the six providers share revenue along the performance frontier (e.g., above a 70\% SWE-bench solve rate). Two models are uncompetitive and earn no revenue: Claude Sonnet 4.5 is too expensive for the budget regimes considered, while Qwen Coder 480B is neither sufficiently cost-efficient nor high-performing.

Lastly, we compare the profitability of arbitrage in the six-model market with that in the previously analyzed two-model market (see Figure~\ref{fig:moremodels}, right). In the six-model market, arbitrage opportunities emerge at a 42\% SWE-bench solve rate, compared to 68\% in the two-model market. Moreover, arbitrage in the six-model market is strictly more profitable, with margins reaching up to 58\%, versus 45\% in the two-model setting. Thus, arbitrage opportunities are both more prevalent and more profitable in the six-model market.

In summary, larger markets are not necessarily more efficient. On the contrary, access to a broader set of providers with varying cost-efficiency can favor arbitrageurs, resulting in arbitrage opportunities across a wider range of performance levels and increased profitability. In the next section, we examine the effectiveness of distillation for training models with different cost–efficiency trade-offs, thereby enabling arbitrage.

\section{Distillation and arbitrage}\label{sec:distill}

Arbitrageurs exploit cost differentials to generate profit opportunities. Model distillation compresses the capabilities of a large teacher into a smaller, cheaper student. In this section, we examine how distillation facilitates arbitrage. First, we show that arbitrage profitability grows monotonically with the distillation budget. Second, we show that distilled models can substantially erode the teacher model's revenue.

\subsection{Distillation creates arbitrage opportunities}

We use Qwen Coder to distill small 1.7B models at different distillation budgets. We then analyze the profitability of arbitrage when pairing each distilled model with Qwen Coder 480B.

To synthesize the training data, we start from SWE-Smith~\citep{yang2025swesmith}, a dataset of over 60k GitHub issues. Approximately 80\% of the issues lack descriptions, which we generate using Qwen 3 235B Instruct. We discard 13\% of issues due to Docker compatibility problems, yielding a final set of 52.4k distinct GitHub issues. We then use Qwen Coder 30B\footnote{We use the 30B model to reduce the cost of data generation. We would expect better performance when using the 480B model.} to generate eight\footnote{Preliminary experiments show that 16 generations per problem underperforms compared to 8 generations per problem.} trajectories per issue, yielding about 400k training trajectories or 5.4B training tokens. Although only 20\% of these trajectories are correct, training on the full set of trajectories results in better downstream performance.

\begin{figure*}[t]
    \centering
    \includegraphics[width=\linewidth]{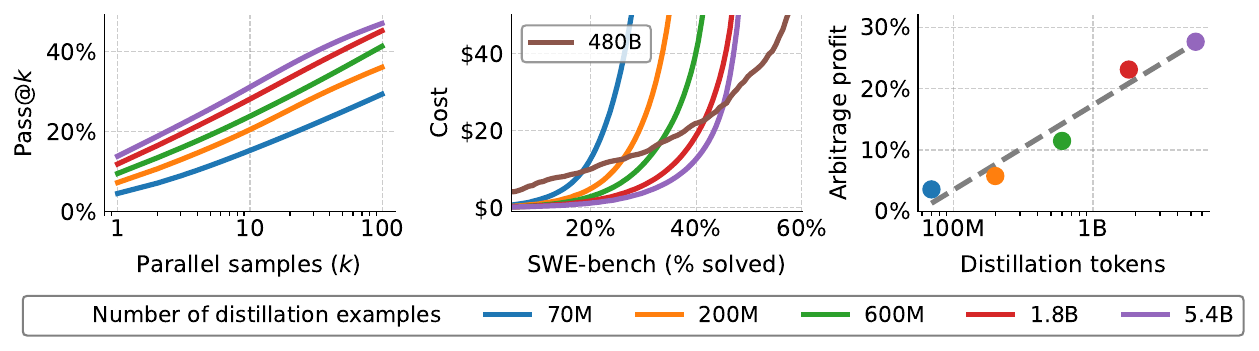}
    \caption{We fine-tune Qwen 3 1.7B using data generated by Qwen Coder 30B. We distill for up to 400k examples (5.4B tokens). \emph{Left}: Models distilled on more data Pareto-dominate in terms of pass@$k$. \emph{Middle}: Models distilled on more data attain higher levels of performance at lower cost. \emph{Right:} When paired with Qwen Coder 480B, models distilled on more data create increasingly more profitable arbitrage opportunities.}
    \label{fig:scaling}
\end{figure*}

We distill five Qwen 3 1.7B models at different data scales: 
70M, 200M, 600M, 1.8B, and 5.4B training tokens
, using standard supervised fine-tuning (SFT). We then evaluate their pass@$k$ performance on SWE-bench, see Figure~\ref{fig:scaling} (left). We find that increased distillation consistently improves pass@$k$. In fact, distillation leads to larger improvements in pass@100 compared to pass@1. 
In turn, models distilled on more data dominate those distilled on less data in terms of their cost-performance, see Figure~\ref{fig:scaling} (middle).

Next, we evaluate the arbitrage opportunities enabled by each distilled model when paired with Qwen Coder 480B. We measure mean profitability on the upper end of performance (61\% to 71\% SWE-bench), and plot mean profitability against the number of distillation tokens in Figure~\ref{fig:scaling} right. We observe that increased distillations consistently create more profitable arbitrage opportunities, with profitability increasing roughly log-linearly with the number of distillation tokens. The model distilled on 5.4B tokens (400k examples) enables a remarkably high level of profitability, allowing for a profit margin of nearly 30\%.

Therefore, distillation is highly effective at creating arbitrage opportunities.  We replicate these scaling experiments in the setting of Lean 4 formal theorem proving and observe consistent findings, see Appendix~\ref{app:leanscaling}. Next, we examine how the teacher model’s revenue changes after distilled models enter the market.

\subsection{Distillation and revenue displacement}

Having established the effectiveness of distillation but lacking additional seed problems to generate more training data, we turn to scaling model size. Specifically, we fine-tune Qwen~3~4B on the full 400k training examples from the previous section and refer to the resulting model as mini-coder 4B. We then examine how the introduction of mini-coder 4B affects the revenue of Qwen Coder 30B in a competitive market. 

\begin{figure*}[t]
    \centering
    \includegraphics[width=\linewidth]{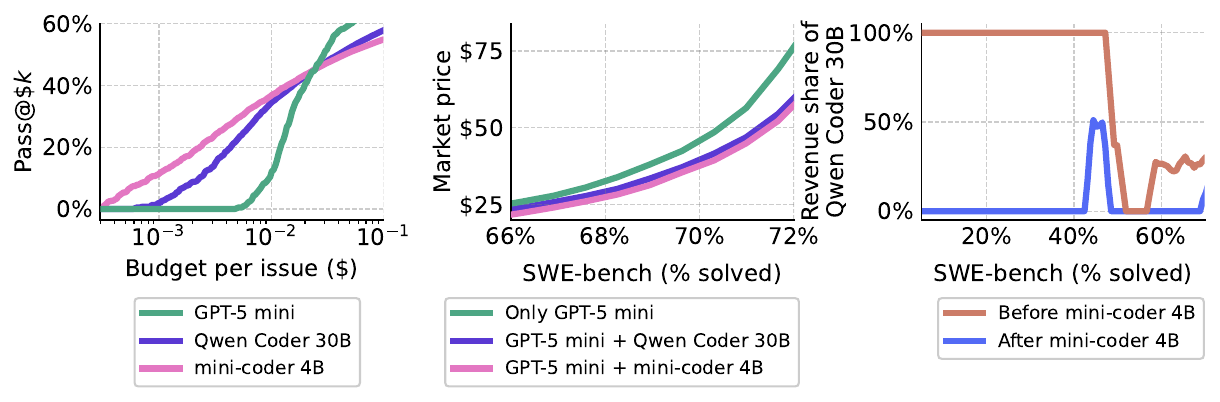}
    \caption{We train mini-coder 4B with data generated by Qwen Coder 30B. \emph{Left:} mini-coder outperforms Qwen Coder at inference budgets up to \$0.02 per issue. \emph{Middle:} When paired with GPT-5 mini, mini-coder leads to lower market prices (i.e., arbitrage buy costs) than Qwen Coder. \emph{Right:} Upon entering a market consisting of GPT-5 mini and Qwen Coder, mini-coder cannibalizes nearly all of Qwen Coder’s revenue.}
    \label{fig:passdistill}
\end{figure*}

To do so, we analyze a three-way market consisting of mini-coder 4B, Qwen Coder 30B, and the more capable GPT-5 mini model. We compare the pass@\$$k$ performance of the three models in Figure~\ref{fig:passdistill} left, where pass@\$$k$ denotes the solve rate under a \$$k$ budget per example. As expected, both mini-coder and Qwen Coder are more cost-efficient than GPT-5 mini at small sampling budgets. However, in this low-budget regime, mini-coder outperforms Qwen Coder; consequently, Qwen Coder is dominated by mini-coder and GPT-5~mini.

To assess whether mini-coder could serve as a replacement for Qwen Coder, we consider two market configurations: GPT-5 mini paired with Qwen Coder, and GPT-5 mini paired with mini-coder. We plot the arbitrage-free market price (i.e., the arbitrage buy cost) in each setting in Figure~\ref{fig:passdistill} middle. The market with mini-coder yields prices comparable to those in the Qwen Coder market, indicating that the distilled model could effectively replace its teacher model without increasing market prices.

We now examine how the entry of mini-coder affects Qwen Coder’s revenue. In Figure~\ref{fig:passdistill} (left), we plot Qwen Coder’s revenue share in two settings: before mini-coder enters the market (i.e., GPT-5 mini + Qwen Coder 30B) and after mini-coder enters (i.e., the three-model market). Before mini-coder’s entry (brown), Qwen Coder captures nearly all revenue in the lower-performance regime and maintains a substantial share of around 20\% at the performance frontier. After mini-coder enters (blue), Qwen Coder’s revenue nearly disappears, with the model earning revenue only within a narrow performance band. In other words, mini-coder cannibalizes nearly all of its teacher model’s revenue.

These results highlight the effectiveness of distillation. Distilled models can outperform their teacher models precisely in the performance regimes that matter most given existing competitors. In doing so, they may cannibalize a large share of the teacher model’s revenue.

\section{Related work}

\paragraph{Pricing model outputs.} Price-per-token pricing is ubiquitous in current model marketplaces. Prior work has identified several potential pitfalls, including token count misrepresentation~\citep{velasco2025auditing, velasco2025your, wang2025predictive, sun2025invisible} and model substitution~\citep{gao2024model, cai2025you, sun2025invisible}. Alternative pricing mechanisms have also been proposed, such as pay-for-performance contracts~\citep{saig2024incentivizing}, second-best performance auctions~\citep{cao2025pay}, and menus of two-part tariffs~\citep{bergemann2025economics}. Our model is agnostic to the specific mechanism used by each provider to price model outputs, provided that cost–performance curves can be computed. Instead, we study the extent to which arbitrageurs can exploit price differentials in the market, which in turn enables us to determine overall market efficiency and the arbitrage-free valuation of different performance levels.

\paragraph{Model cascading and routing.} Model cascading sequentially queries multiple models, typically in increasing order of cost, until a response satisfying a predefined quality criterion is obtained~\citep{varshney2022model, wang2023tabi, madaan2023automix, chenfrugalgpt, ramirezoptimising, zhang2024ecoassistant, kapoor2025ai}. Model routing, by contrast, assigns different queries to different models in order to maximize performance, minimize cost, or a combination of both~\citep{shnitzerlarge, dinghybrid, lu2024routing, vsakota2024fly}. Arbitrageurs may draw upon the literature on model cascading and model routing to construct arbitrage policies. We adopt a cascade-style design, scaling inference-time compute across the cascade, and optimizing the inference budget allocation to maximize arbitrage profit. Conversely, research on model routing and cascading can adopt arbitrage profitability as key benchmark for algorithm development, with improvements in profitability translating into gains in market efficiency.

\paragraph{Inference-time scaling.} Inference-time scaling allows model performance to be traded off against inference cost. We scale inference-time compute through repeated sampling~\citep{gsm8k, humaneval, li2022competition}. Other approaches could also be considered, such as test-time search~\citep{yao2023tree} or test-time training~\citep{hardttest}. Meaningful comparisons across models and inference-time strategies require a standardized measure of inference cost. The literature typically uses floating-point operations (FLOPs) for this purpose~\citep{hassidlarger, wu2025inference, brown2025large}. Instead, we measure inference cost in USD using OpenRouter’s prices, which more directly reflect usage costs. More importantly, our arbitrage framework offers a novel way to benchmark different combinations of models and inference-time scaling strategies by evaluating how much revenue they can generate in a competitive market.

\paragraph{Distillation.} We examine how test-time scaling capabilities transfer through distillation. While modest amounts of distillation can improve pass@$k$~\citep{humaneval, yue2025does}, excessive fine-tuning may lead to diversity collapse and reduced pass@$k$ performance~\citep{gsm8k, chen2025rethinking, dang2025weight}. To mitigate this issue, inference-aware fine-tuning methods have been proposed~\citep{chowinference, chen2025rethinking, goyal2025distilled, dang2025weight}. In contrast, we distill models using standard supervised fine-tuning with the cross-entropy loss. We find that test-time scaling performance (e.g., pass@100) continues to improve as the number of distillation tokens scales into the billions.

More broadly, distillation typically produces models that are less capable but cheaper to run. This makes it difficult to assess their value, given that a more powerful teacher model is necessarily also available.
Our arbitrage framework allows us to quantify the economic value of distilled models by measuring the revenue they can generate upon market entry, as well as the extent to which they drive down market prices.

\paragraph{Verification.} For some tasks, verifying the correctness of a solution is easier than generating correct solutions~\citep{gsm8k}. The more verifiable a task is, the easier it tends to be to improve model performance~\citep{wei2025asymmetry, keles2025verifiability}. For example, improvements can be achieved at training time through RL with verifiable rewards~\citep{guo2025deepseek}, or at test time by scaling inference compute~\citep{humaneval}.

Our work highlights an additional consequence of verifiability: the emergence of AI model markets. When model outputs are verifiable, customers can choose freely among different providers without requiring prior familiarity or trust. Verifiable solutions become fungible goods and are therefore subject to classical economic analysis. As a result, verification facilitates not only higher-performing models but also enables the development of competitive model markets. Although verification may entail costs~\citep{gdpval}, these costs can be incorporated straightforwardly into our economic analysis.

\paragraph{Economic analyses of AI model ecosystems.} Prior work examines various economic aspects of model ecosystems. \citet{erol2025cost} analyze differences in the expected cost of producing a correct output across language models. In contrast, we examine how arbitrageurs can exploit these cost differentials and study the resulting economic implications. \citet{bergemann2025economics} develop an economic framework for optimal pricing of model inference and fine-tuning; we instead focus on the arbitrage-free valuation of model performance. \citet{xu2025economics} investigate how different economic factors shape model providers’ openness decisions; arbitrage may incentivize dominant providers to gatekeep small, efficient models. Finally, \citet{jagadeesan2025safety} show that multi-objective model development lowers barriers to market entry. We focus on cost-efficiency rather than safety, and demonstrate how arbitrage facilitates market entry.

\section{Discussion}

We initiate the study of computational arbitrage in AI model markets. We demonstrate its feasibility through a case study on SWE-bench GitHub issue resolution, and analyze several economic implications, such as reductions in market prices and lowered barriers to entry. We then study the interplay between computational arbitrage and model distillation: the existence of arbitrageurs incentivizes the market entry of distilled models, which in turn create the very opportunities these arbitrageurs exploit.

Many theoretical and empirical questions remain open. In this work we assume verification to be costless. An important research avenue is to investigate the implications of costly or imperfect verification~\citep{gdpval}. We have also assumed perfect market information. The cost of market information~\citep{stigler1962information} may determine whether computational arbitrage meaningfully improves market efficiency or merely shifts market concentration from model providers toward oligopolistic intermediaries.

From a technical perspective, the arbitrage strategies we consider are both query‑agnostic and user‑agnostic. While already remarkably effective, practitioners may draw on insights from the rich literature on model routing to devise more sophisticated arbitrage policies. Active learning approaches could allow arbitrageurs to dynamically adapt to evolving query distributions and market conditions. These technical improvements, insofar as they promote market efficiency, stand to benefit consumers and providers alike; the former through lower prices, the latter through lowered barriers to entry.

\bibliography{bib}

@article{humaneval,
  title={Evaluating large language models trained on code},
  author={Chen, Mark and Tworek, Jerry and Jun, Heewoo and Yuan, Qiming and Pinto, Henrique Ponde De Oliveira and Kaplan, Jared and Edwards, Harri and Burda, Yuri and Joseph, Nicholas and Brockman, Greg and others},
  journal={arXiv preprint arXiv:2107.03374},
  year={2021}
}

@article{gsm8k,
  title={Training verifiers to solve math word problems},
  author={Cobbe, Karl and Kosaraju, Vineet and Bavarian, Mohammad and Chen, Mark and Jun, Heewoo and Kaiser, Lukasz and Plappert, Matthias and Tworek, Jerry and Hilton, Jacob and Nakano, Reiichiro and others},
  journal={arXiv preprint arXiv:2110.14168},
  year={2021}
}

@inproceedings{moura2021lean,
  title={The lean 4 theorem prover and programming language},
  author={Moura, Leonardo de and Ullrich, Sebastian},
  booktitle={International Conference on Automated Deduction},
  pages={625--635},
  year={2021},
  organization={Springer}
}

@inproceedings{hoffmann2022training,
  title={Training compute-optimal large language models},
  author={Hoffmann, Jordan and Borgeaud, Sebastian and Mensch, Arthur and Buchatskaya, Elena and Cai, Trevor and Rutherford, Eliza and de Las Casas, Diego and Hendricks, Lisa Anne and Welbl, Johannes and Clark, Aidan and others},
  booktitle={Proceedings of the 36th International Conference on Neural Information Processing Systems},
  pages={30016--30030},
  year={2022}
}

@misc{sonnet,
  title       = {Claude Sonnet 4.5 System Card},
  author      = {{Anthropic}},
  year        = {2025},
  url         = {https://www-cdn.anthropic.com/963373e433e489a87a10c823c52a0a013e9172dd.pdf},
}

@inproceedings{minif2f,
  title={miniF2F: a cross-system benchmark for formal Olympiad-level mathematics},
  author={Zheng, Kunhao and Han, Jesse Michael and Polu, Stanislas},
  booktitle={International Conference on Learning Representations},
  year={2022},
}

@misc{
brown2025large,
title={Large Language Monkeys: Scaling Inference Compute with Repeated Sampling},
author={Bradley Brown and Jordan Juravsky and Ryan Saul Ehrlich and Ronald Clark and Quoc V Le and Christopher Re and Azalia Mirhoseini},
year={2025},
url={https://openreview.net/forum?id=0xUEBQV54B}
}

@article{li2022competition,
  title={Competition-level code generation with alphacode},
  author={Li, Yujia and Choi, David and Chung, Junyoung and Kushman, Nate and Schrittwieser, Julian and Leblond, R{\'e}mi and Eccles, Tom and Keeling, James and Gimeno, Felix and Dal Lago, Agustin and others},
  journal={Science},
  volume={378},
  number={6624},
  pages={1092--1097},
  year={2022},
  publisher={American Association for the Advancement of Science}
}

@inproceedings{hassidlarger,
  title={The Larger the Better? Improved LLM Code-Generation via Budget Reallocation},
  author={Hassid, Michael and Remez, Tal and Gehring, Jonas and Schwartz, Roy and Adi, Yossi},
  booktitle={First Conference on Language Modeling},
  year={2024}
}

@inproceedings{
    swebench,
    title={{SWE}-bench: Can Language Models Resolve Real-world Github Issues?},
    author={Carlos E Jimenez and John Yang and Alexander Wettig and Shunyu Yao and Kexin Pei and Ofir Press and Karthik R Narasimhan},
    booktitle={The Twelfth International Conference on Learning Representations},
    year={2024},
    url={https://openreview.net/forum?id=VTF8yNQM66}
}

@misc{verified,
  title={Introducing {SWE}-bench Verified},
  author={Chowdhury, Neil and Aung, James and Shern, Chan Jun and Jaffe, Oliver and Sherburn, Dane and Starace, Giulio and Mays, Evan and Dias, Rachel and Aljubeh, Marwan and Glaese, Mia and Jimenez, Carlos E. and Yang, John and Ho, Leyton and Patwardhan, Tejal and Liu, Kevin and Madry, Aleksander},
  year={2024},
  url={https://openai.com/index/introducing-swe-bench-verified/},
}

@article{wang2025kimina,
  title={Kimina-Prover Preview: Towards Large Formal Reasoning Models with Reinforcement Learning},
  author={Wang, Haiming and Unsal, Mert and Lin, Xiaohan and Baksys, Mantas and Liu, Junqi and Dos Santos, Marco and Sung, Flood and Vinyes, Marina and Ying, Zhenzhe and Zhu, Zekai and others},
  journal={CoRR},
  year={2025}
}

@misc{yang2025qwen3,
      title={Qwen3 Technical Report}, 
      author={Qwen Team},
      year={2025},
      eprint={2505.09388},
      archivePrefix={arXiv},
      primaryClass={cs.CL},
      url={https://arxiv.org/abs/2505.09388},
}

@article{qwencoder,
  title={Qwen2. 5-Coder Technical Report},
  author={Hui, Binyuan and Yang, Jian and Cui, Zeyu and Yang, Jiaxi and Liu, Dayiheng and Zhang, Lei and Liu, Tianyu and Zhang, Jiajun and Yu, Bowen and Dang, Kai and others},
  journal={arXiv preprint arXiv:2409.12186},
  year={2024}
}

@misc{numinamath,
  author = {Jia LI and Edward Beeching and Lewis Tunstall and Ben Lipkin and Roman Soletskyi and Shengyi Costa Huang and Kashif Rasul and Longhui Yu and Albert Jiang and Ziju Shen and Zihan Qin and Bin Dong and Li Zhou and Yann Fleureau and Guillaume Lample and Stanislas Polu},
  title = {NuminaMath},
  year = {2024},
  publisher = {Numina},
  journal = {Hugging Face repository},
  howpublished = {\url{[https://huggingface.co/AI-MO/NuminaMath-1.5](https://github.com/project-numina/aimo-progress-prize/blob/main/report/numina_dataset.pdf)}}
}

@misc{yang2025swesmith,
  title={SWE-smith: Scaling Data for Software Engineering Agents}, 
  author={John Yang and Kilian Lieret and Carlos E. Jimenez and Alexander Wettig and Kabir Khandpur and Yanzhe Zhang and Binyuan Hui and Ofir Press and Ludwig Schmidt and Diyi Yang},
  year={2025},
  eprint={2504.21798},
  archivePrefix={arXiv},
  primaryClass={cs.SE},
  url={https://arxiv.org/abs/2504.21798},
}

@inproceedings{yang2024sweagent,
  title={{SWE}-agent: Agent-Computer Interfaces Enable Automated Software Engineering},
  author={John Yang and Carlos E Jimenez and Alexander Wettig and Kilian Lieret and Shunyu Yao and Karthik R Narasimhan and Ofir Press},
  booktitle={The Thirty-eighth Annual Conference on Neural Information Processing Systems},
  year={2024},
  url={https://arxiv.org/abs/2405.15793}
}

@article{chen2025rethinking,
  title={Rethinking fine-tuning when scaling test-time compute: Limiting confidence improves mathematical reasoning},
  author={Chen, Feng and Raventos, Allan and Cheng, Nan and Ganguli, Surya and Druckmann, Shaul},
  journal={arXiv preprint arXiv:2502.07154},
  year={2025}
}

@article{dang2025weight,
  title={Weight ensembling improves reasoning in language models},
  author={Dang, Xingyu and Baek, Christina and Wen, Kaiyue and Kolter, Zico and Raghunathan, Aditi},
  journal={arXiv preprint arXiv:2504.10478},
  year={2025}
}

@inproceedings{yue2025does,
  title={Does Reinforcement Learning Really Incentivize Reasoning Capacity in LLMs Beyond the Base Model?},
  author={Yue, Yang and Chen, Zhiqi and Lu, Rui and Zhao, Andrew and Wang, Zhaokai and Song, Shiji and Huang, Gao},
  booktitle={The Thirty-ninth Annual Conference on Neural Information Processing Systems},
  year={2025}
}

@misc{keles2025verifiability,
  author= {Alperen Keles},
  title= {Verifiability is the Limit},
  year= {2025},
  howpublished={https://alperenkeles.com/posts/verifiability-is-the-limit/},
  note={Accessed: 2026-03-01}
}

@inproceedings{chowinference,
  title={Inference-Aware Fine-Tuning for Best-of-N Sampling in Large Language Models},
  author={Chow, Yinlam and Tennenholtz, Guy and Gur, Izzeddin and Zhuang, Vincent and Dai, Bo and Kumar, Aviral and Agarwal, Rishabh and Thiagarajan, Sridhar and Boutilier, Craig and Faust, Aleksandra},
  booktitle={The Thirteenth International Conference on Learning Representations},
  year={2025}
}

@article{goyal2025distilled,
  title={Distilled Pretraining: A modern lens of Data, In-Context Learning and Test-Time Scaling},
  author={Goyal, Sachin and Lopez-Paz, David and Ahuja, Kartik},
  journal={arXiv preprint arXiv:2509.01649},
  year={2025}
}

@inproceedings{wu2025inference,
  title={Inference scaling laws: An empirical analysis of compute-optimal inference for LLM problem-solving},
  author={Wu, Yangzhen and Sun, Zhiqing and Li, Shanda and Welleck, Sean and Yang, Yiming},
  booktitle={The Thirteenth International Conference on Learning Representations},
  year={2025}
}

@article{guo2025deepseek,
  title={Deepseek-r1: Incentivizing reasoning capability in llms via reinforcement learning},
  author={Guo, Daya and Yang, Dejian and Zhang, Haowei and Song, Junxiao and Zhang, Ruoyu and Xu, Runxin and Zhu, Qihao and Ma, Shirong and Wang, Peiyi and Bi, Xiao and others},
  journal={arXiv preprint arXiv:2501.12948},
  year={2025}
}

@misc{wei2025asymmetry,
  author       = {Jason Wei},
  title        = {Asymmetry of verification and verifier’s rule},
  year         = {2025},
  howpublished          = {https://www.jasonwei.net/blog/asymmetry-of-verification-and-verifiers-law},
}

@article{chenfrugalgpt,
  title={FrugalGPT: How to Use Large Language Models While Reducing Cost and Improving Performance},
  author={Chen, Lingjiao and Zaharia, Matei and Zou, James},
  journal={Transactions on Machine Learning Research},
  year={2023}
}

@inproceedings{shnitzerlarge,
  title={Large Language Model Routing with Benchmark Datasets},
  author={Shnitzer, Tal and Ou, Anthony and Silva, M{\'\i}rian and Soule, Kate and Sun, Yuekai and Solomon, Justin and Thompson, Neil and Yurochkin, Mikhail},
  booktitle={First Conference on Language Modeling},
  year={2024}
}

@inproceedings{vsakota2024fly,
  title={Fly-swat or cannon? cost-effective language model choice via meta-modeling},
  author={{\v{S}}akota, Marija and Peyrard, Maxime and West, Robert},
  booktitle={Proceedings of the 17th ACM International Conference on Web Search and Data Mining},
  pages={606--615},
  year={2024}
}

@inproceedings{lu2024routing,
  title={Routing to the Expert: Efficient Reward-guided Ensemble of Large Language Models},
  author={Lu, Keming and Yuan, Hongyi and Lin, Runji and Lin, Junyang and Yuan, Zheng and Zhou, Chang and Zhou, Jingren},
  booktitle={Proceedings of the 2024 Conference of the North American Chapter of the Association for Computational Linguistics: Human Language Technologies (Volume 1: Long Papers)},
  pages={1964--1974},
  year={2024}
}

@inproceedings{wang2023tabi,
  title={Tabi: An efficient multi-level inference system for large language models},
  author={Wang, Yiding and Chen, Kai and Tan, Haisheng and Guo, Kun},
  booktitle={Proceedings of the Eighteenth European Conference on Computer Systems},
  pages={233--248},
  year={2023}
}

@article{madaan2023automix,
  title={AutoMix: Automatically Mixing Language Models},
  author={Madaan, Aman and Aggarwal, Pranjal and Anand, Ankit and Potharaju, Srividya Pranavi and Mishra, Swaroop and Zhou, Pei and Gupta, Aditya and Rajagopal, Dheeraj and Kappaganthu, Karthik and Yang, Yiming and others},
  journal={CoRR},
  year={2023}
}

@inproceedings{ramirezoptimising,
  title={Optimising Calls to Large Language Models with Uncertainty-Based Two-Tier Selection},
  author={Ram{\'\i}rez, Guillem and Birch, Alexandra and Titov, Ivan},
  booktitle={First Conference on Language Modeling},
  year={2024}
}

@inproceedings{dinghybrid,
  title={Hybrid LLM: Cost-Efficient and Quality-Aware Query Routing},
  author={Ding, Dujian and Mallick, Ankur and Wang, Chi and Sim, Robert and Mukherjee, Subhabrata and R{\"u}hle, Victor and Lakshmanan, Laks VS and Awadallah, Ahmed Hassan},
  booktitle={The Twelfth International Conference on Learning Representations},
  year={2024}
}

@inproceedings{varshney2022model,
  title={Model Cascading: Towards Jointly Improving Efficiency and Accuracy of NLP Systems},
  author={Varshney, Neeraj and Baral, Chitta},
  booktitle={Proceedings of the 2022 Conference on Empirical Methods in Natural Language Processing},
  pages={11007--11021},
  year={2022}
}

@inproceedings{zhang2024ecoassistant,
  title={EcoAssistant: Using LLM Assistants More Affordably and Accurately},
  author={Zhang, Jieyu and Krishna, Ranjay and Awadallah, Ahmed Hassan and Wang, Chi},
  booktitle={ICLR 2024 Workshop on Large Language Model (LLM) Agents},
  year={2024}
}

@article{stigler1962information,
  title={Information in the labor market},
  author={Stigler, George J},
  journal={Journal of political economy},
  volume={70},
  number={5, Part 2},
  pages={94--105},
  year={1962},
  publisher={The University of Chicago Press}
}

@article{velasco2025auditing,
  title={Auditing Pay-Per-Token in Large Language Models},
  author={Velasco, Ander Artola and Tsirtsis, Stratis and Gomez-Rodriguez, Manuel},
  journal={arXiv preprint arXiv:2510.05181},
  year={2025}
}

@article{cai2025you,
  title={Are you getting what you pay for? auditing model substitution in llm apis},
  author={Cai, Will and Shi, Tianneng and Zhao, Xuandong and Song, Dawn},
  journal={arXiv preprint arXiv:2504.04715},
  year={2025}
}

@article{xu2025economics,
  title={The economics of AI foundation models: Openness, competition, and governance},
  author={Xu, Fasheng and Wang, Xiaoyu and Chen, Wei and Xie, Karen},
  journal={arXiv preprint arXiv:2510.15200},
  year={2025}
}

@article{wang2025predictive,
  title={Predictive Auditing of Hidden Tokens in LLM APIs via Reasoning Length Estimation},
  author={Wang, Ziyao and Sun, Guoheng and He, Yexiao and Shen, Zheyu and Tian, Bowei and Li, Ang},
  journal={arXiv preprint arXiv:2508.00912},
  year={2025}
}

@article{sun2025invisible,
  title={Invisible Tokens, Visible Bills: The Urgent Need to Audit Hidden Operations in Opaque LLM Services},
  author={Sun, Guoheng and Wang, Ziyao and Zhao, Xuandong and Tian, Bowei and Shen, Zheyu and He, Yexiao and Xing, Jinming and Li, Ang},
  journal={arXiv preprint arXiv:2505.18471},
  year={2025}
}

@article{velasco2025your,
  title={Is Your LLM Overcharging You? Tokenization, Transparency, and Incentives},
  author={Velasco, Ander Artola and Tsirtsis, Stratis and Okati, Nastaran and Gomez-Rodriguez, Manuel},
  journal={arXiv preprint arXiv:2505.21627},
  year={2025}
}

@article{erol2025cost,
  title={Cost-of-Pass: An Economic Framework for Evaluating Language Models},
  author={Erol, Mehmet Hamza and El, Batu and Suzgun, Mirac and Yuksekgonul, Mert and Zou, James},
  journal={arXiv preprint arXiv:2504.13359},
  year={2025}
}

@article{
kapoor2025ai,
title={{AI} Agents That Matter},
author={Sayash Kapoor and Benedikt Stroebl and Zachary S Siegel and Nitya Nadgir and Arvind Narayanan},
journal={Transactions on Machine Learning Research},
issn={2835-8856},
year={2025},
url={https://openreview.net/forum?id=Zy4uFzMviZ},
note={}
}

@inproceedings{bergemann2025economics,
  title={The Economics of Large Language Models: Token Allocation, Fine-Tuning, and Optimal Pricing},
  author={Bergemann, Dirk and Bonatti, Alessandro and Smolin, Alex},
  booktitle={Proceedings of the 26th ACM Conference on Economics and Computation},
  pages={786--786},
  year={2025}
}

@article{cao2025pay,
  title={Pay for The Second-Best Service: A Game-Theoretic Approach Against Dishonest LLM Providers},
  author={Cao, Yuhan and Wang, Yu and Liu, Sitong and Li, Miao and Tao, Yixin and He, Tianxing},
  journal={arXiv preprint arXiv:2511.00847},
  year={2025}
}

@article{saig2024incentivizing,
  title={Incentivizing quality text generation via statistical contracts},
  author={Saig, Eden and Einav, Ohad and Talgam-Cohen, Inbal},
  journal={Advances in Neural Information Processing Systems},
  volume={37},
  pages={51196--51222},
  year={2024}
}

@article{gao2024model,
  title={Model Equality Testing: Which Model Is This API Serving?},
  author={Gao, Irena and Liang, Percy and Guestrin, Carlos},
  journal={ICLR},
  year={2025}
}

@book{mas1995microeconomic,
  title={Microeconomic theory},
  author={Mas-Colell, Andreu and Whinston, Michael Dennis and Green, Jerry R and others},
  volume={1},
  year={1995},
  publisher={Oxford university press New York}
}

@book{hulloptions,
	author = {Hull, John, 1946- author.},
	title = {Options, futures, and other derivative securities },
	publisher = {Prentice Hall},
	year = {1989},
	address = {United states of America},
	edition = {First edition}
}

@TechReport{limitsarbitrage,
type={NBER Working Papers},
institution={National Bureau of Economic Research, Inc},
author={Andrei Shleifer and Robert W. Vishny},
title={The Limits of Arbitrage},
year={1995},
month={Jul},
number={5167},
doi={None},
}

@misc{chernova2025openrouter,
  author       = {Chernova, Yuliya},
  title        = {OpenRouter, a Marketplace for AI Models, Raises \$40 Million},
  howpublished = {The Wall Street Journal},
  year         = {2025},
  month        = jun,
  day          = {25},
  urldate      = {2026-01-28}
}

@article{jagadeesan2025safety,
  title={Safety versus performance: How multi-objective learning reduces barriers to market entry},
  author={Jagadeesan, Meena and Jordan, Michael I and Steinhardt, Jacob},
  journal={Proceedings of the National Academy of Sciences},
  volume={122},
  number={42},
  pages={e2510004122},
  year={2025},
  publisher={National Academy of Sciences}
}

@article{gdpval,
  title={GDPval: Evaluating AI Model Performance on Real-World Economically Valuable Tasks},
  author={Tejal Patwardhan and Rachel Dias and Elizabeth Proehl and Grace Kim and Michele Wang and Olivia Watkins and Sim'on Posada Fishman and Marwan Aljubeh and Phoebe Thacker and Laurance Fauconnet and Natalie S. Kim and Patrick Chao and Samuel Miserendino and Gildas Chabot and David Li and Michael Sharman and Alexandra Barr and Amelia Glaese and Jerry Tworek},
  journal={ArXiv},
  year={2025},
  volume={abs/2510.04374},
}

@article{singh2025openai,
  title={Openai gpt-5 system card},
  author={Singh, Aaditya and Fry, Adam and Perelman, Adam and Tart, Adam and Ganesh, Adi and El-Kishky, Ahmed and McLaughlin, Aidan and Low, Aiden and Ostrow, AJ and Ananthram, Akhila and others},
  journal={arXiv preprint arXiv:2601.03267},
  year={2025}
}

@article{liu2025deepseek,
  title={Deepseek-v3. 2: Pushing the frontier of open large language models},
  author={Liu, Aixin and Mei, Aoxue and Lin, Bangcai and Xue, Bing and Wang, Bingxuan and Xu, Bingzheng and Wu, Bochao and Zhang, Bowei and Lin, Chaofan and Dong, Chen and others},
  journal={arXiv preprint arXiv:2512.02556},
  year={2025}
}

@article{yao2023tree,
  title={Tree of thoughts: Deliberate problem solving with large language models},
  author={Yao, Shunyu and Yu, Dian and Zhao, Jeffrey and Shafran, Izhak and Griffiths, Tom and Cao, Yuan and Narasimhan, Karthik},
  journal={Advances in neural information processing systems},
  volume={36},
  pages={11809--11822},
  year={2023}
}

@inproceedings{hardttest,
  title={Test-Time Training on Nearest Neighbors for Large Language Models},
  author={Hardt, Moritz and Sun, Yu},
  booktitle={The Twelfth International Conference on Learning Representations},
  year={2024}
}

\newpage
\appendix

\section{Evaluation details}
\label{app:eval}

We evaluate on a subset of 445 problems from SWE-bench Verified, rather than the full set of 500 issues (i.e., 89\% of the benchmark). This restriction is due to incompatibility issues between our local cluster and several of the SWE-bench Docker images, which prevent certain instances from running successfully. These incompatibilities primarily affect issues originating from the matplotlib repository.

For generation, we use the lightweight mini-coder-v1 scaffolding~\citep{yang2024sweagent}, which enables models to interact with the Docker environment via bash commands. We make a minor modification to this scaffolding by truncating each model response after the first bash command (i.e., the first bash “block”) instead of returning an error message. Unless otherwise specified, we sample with a temperature of 0.6 and use each model’s default sampling parameters; the exception is GPT-5 mini, which does not expose a temperature parameter. For reasoning models that support a reasoning-effort parameter, we set this parameter to “medium”. We allow for a maximum generation budget of 250 turns, or \$1 in inference budget.

\section{Pricing details}
\label{app:price}

We evaluate some models via API queries and others in our local computing cluster. We log the number of input and output tokens and compute the cost in USD using OpenRouter pricing as of January 2025. We use a 90\% price reduction for cached inputs, in line with the pricing policies of the OpenAI, Claude, and DeepSeek API platforms. Table~\ref{tab:prices} summarizes the price-per-token values used.

For mini-coder 4B, we adopt the pricing of Gemma 3 4B, a similarly sized model. Because no models comparable in size to mini-coder 1.7B were available on OpenRouter as of January 2025, we estimate its cost as 40\% of mini-coder 4B, approximately matching the ratio of their parameter counts. 

\begin{table}[h]
\centering
\caption{Model pricing per 1M tokens.}
\label{tab:prices}
\begin{tabular}{lccc}
\toprule
\textbf{Model} & \textbf{Input (\$)} & \textbf{Output (\$)} & \textbf{Cache Reduction} \\
\midrule
Claude Sonnet 4.5 & 3.00  & 15.00  & 90\% \\
GPT-5 mini & 0.25  & 2.00  & 90\% \\
Qwen 3 Coder 480B & 0.25  & 1.00  & 90\% \\
DeepSeek v3.2 Reasoner & 0.28  & 0.42  & 90\% \\
Qwen 3 Coder 30B  & 0.07  & 0.27  & 90\% \\
mini-coder 4B     & 0.02  & 0.07  & 90\% \\
mini-coder 1.7B   & 0.008 & 0.028 & 90\% \\
\bottomrule
\end{tabular}
\end{table}

\subsection{Pricing corrections}

\begin{figure*}[t]
    \centering
    \includegraphics[width=\linewidth]{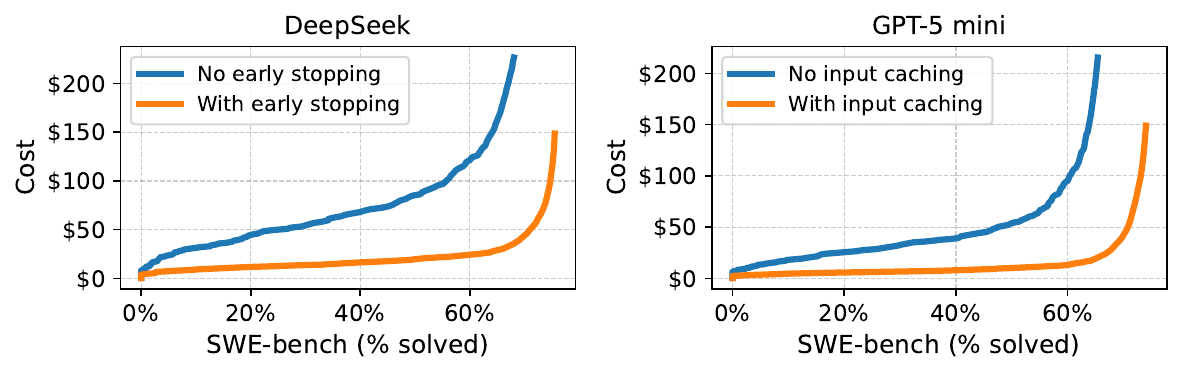}
    \caption{\emph{Left:} price correction by enforcing early-stopping after the first bash command is generated. \emph{Right}: price correction by applying the adverstised 90\% caching discount to GPT-5 mini.}
    \label{fig:pricecorr}
\end{figure*}

\paragraph{Early stopping generations.}
The mini-swe-agent v1 scaffolding expects a single bash command per turn, but models sometimes fail to follow this template. We therefore modify the scaffolding to truncate each model response after the first bash command (i.e., the first bash “block”). For simplicity, we allow models to generate their full response and apply this truncation post hoc. However, a more cost-effective approach would enforce the early-stopping criterion during generation. Accordingly, we price trajectories by counting only the tokens up to the first bash block. This adjustment can substantially affect pricing estimates; see DeepSeek V3.2 Reasoner for an example in Figure~\ref{fig:pricecorr} left.

\paragraph{GPT-5 and caching.} We identified a bug in which GPT-5 mini does not cache inputs across turns, resulting in no cost reduction for previously processed context during multi-turn evaluation. We regard this as a bug in the OpenAI API platform rather than intended behavior. To account for this issue, we adjust the trajectory costs by applying the advertised 90\% caching discount. Figure~\ref{fig:pricecorr} right illustrates the impact of the caching bug on the cost–performance trade-off of GPT-5 mini.

\section{Lean4 theorem proving}
\label{app:lean}
We conduct experiments similar to those in the main text, but for formal theorem proving rather than software issue resolution. Specifically, we consider formal theorem proving using Lean 4~\citep{moura2021lean}, a programming language and proof assistant for formal mathematics. This domain is verifiable, as the Lean 4 compiler can determine whether a given proof correctly proves a given input statement.

We use the MiniF2F benchmark \citep{minif2f} for evaluation, which includes mathematical problems drawn from high-school, undergraduate, and olympiad exercises. We evaluate the Kimina Prover family of models \citep{wang2025kimina}, with model sizes of 0.6B, 1.7B, 8B, and 72B. We scale inference compute via repeated sampling. In contrast to the experiments in the main text, we measure inference budget in floating-point operations (FLOPs), since there are no competitive offerings for the Qwen 3 models on OpenRouter, making comparisons in terms of USD difficult. We approximate inference FLOPs as $C = 2 \times N \times D$ \citep{hoffmann2022training}, where $N$ is the model size and $D$ is the number of generated tokens. We scale inference compute for budgets of up to $3\times10^8$ FLOPs per input problem.

\begin{figure*}[t]
    \centering
    \includegraphics[width=\linewidth]{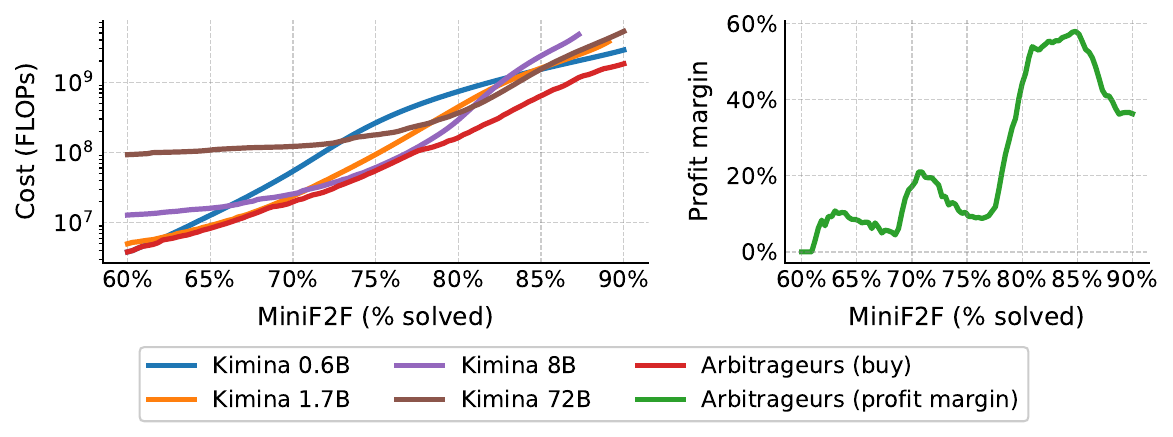}
    \caption{\emph{Left}: Inference cost for each Kimina Prover model to reach various MiniF2F solve rates. We plot in red the best inference cost achievable by an arbitrageur that distributes inference compute across the Kimina models. \emph{Right}: Arbitrageurs can create opportunities for profit, achieving over 60\% profit margin.}
    \label{fig:lean-main}
\end{figure*}

We next examine the revenue earned by each model in the market. For each performance level between 45\% and 75\% SWE-bench solve rate, we search for the arbitrage policy with the lowest purchase cost. This cost represents the arbitrage-free (i.e., equilibrium) market price for that level of performance. We plot the corresponding revenue shares in Figure~\ref{fig:moremodels} (middle). The arbitrage-free market is not segmented: four of the six providers share revenue along the performance frontier (e.g., above a 70\% SWE-bench solve rate). Two models are uncompetitive and earn no revenue: Claude Sonnet 4.5 is too expensive for the budget regimes considered, while Qwen Coder 480B is neither sufficiently cost-efficient nor high-performing.
We plot the cost–performance curves for each of the Kimina models in Figure~\ref{fig:lean-main} (left). We then compute arbitrage profitability as follows. For each MiniF2F performance level between 60\% and 92\% solve rate, we search for the arbitrage policy with the lowest purchase cost, plotted in red. We compute arbitrage profitability by comparing this arbitrage buy cost with the minimum cost for the corresponding performance level across the Kimina models. We plot arbitrage profitability in Figure~\ref{fig:lean-main} (right). Arbitrage opportunities exist for all performance levels above a 61\% solve rate, with remarkably large profit margins of up to 60\%.

\begin{figure*}[t]
    \centering
    \includegraphics[width=\linewidth]{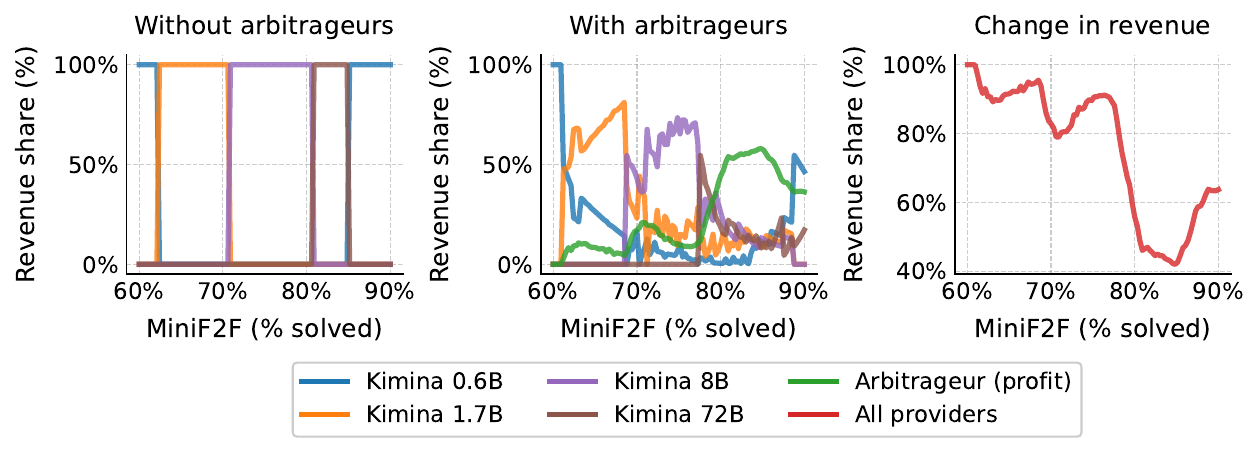}
    \caption{MiniF2F theorem proving. Revenue split across model providers for different levels of performance. \emph{Left:} In the absence of arbitrageurs, the market is segmented by performance, with a single provider dominating each segment. \emph{Middle:} Arbitrageurs eliminate this segmentation, allowing providers to earn revenue across broader ranges of performance. \emph{Right:} Arbitrageurs reduce providers’ marginal revenue.}
    \label{fig:lean-split}
\end{figure*}

We further plot revenue share across providers in Figure~\ref{fig:lean-split}. In the absence of arbitrageurs, the market is segmented, with a single provider dominating each market segment. In contrast, in the presence of arbitrageurs, providers earn revenue across much broader performance levels. For example, between 78\% and 88\% solve rate, all four Kimina models earn revenue. As demonstrated earlier, arbitrage profits (or reduction in market prices) come from reductions in providers' revenue, with the overall marginal revenue of the Kimina models reducing by up to 60\%.

\subsection{Distillation experiments}
\label{app:leanscaling}

\begin{figure*}[t]
    \centering
    \includegraphics[width=\linewidth]{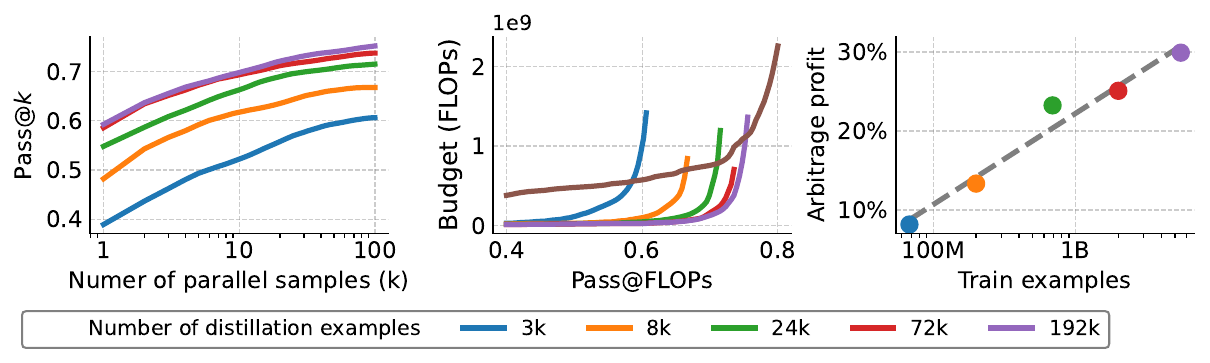}
    \caption{Scaling experiments for Lean theorem proving. We fine-tune Qwen 3 1.7B using synthetic training trajectories from Kimina Prover 1.7B. We distill for up to 200k examples (5B tokens). \emph{Left}: Increased distillation improves test-time scaling. Models distilled on more data Pareto-dominate in terms of pass@$k$. \emph{Middle}
    : We explicitly consider inference budget in FLOPs (i.e., pass@FLOPs). We similarly find that increased distillation monotonically improves pass@FLOPs. \emph{Right:} Arbitrage profit when pairing each of the distilled models with Kimina Prover 72B. Models distilled on more teacher data create increasingly more profitable arbitrage opportunities.}
    \label{fig:lean-scale}
\end{figure*}

We use Kimina Prover 1.7B as the teacher model to reduce the cost of generating training data. We expect that using Kimina 72B for generation would yield stronger arbitrage results. We use Qwen 3 1.7B \citep{yang2025qwen3} as the student model. For the seed problems used to generate training trajectories, we use NuminaMath-LEAN \citep{wang2025kimina, numinamath}, which contains 104,000 mathematical competition problems formalized in Lean 4. We sample 8 teacher responses per problem, yielding 832k synthetic responses for NuminaMath-LEAN.
We distill models on 68M, 207M, 690M, 2B, and 5.5B tokens. We plot the results in Figure~\ref{fig:lean-scale}. We find that increased distillation consistently improves pass@kk
k, with larger improvements in pass@100 than in pass@1.

In turn, models distilled on more data dominate those distilled on less data in terms of cost–performance (see Figure~\ref{fig:scaling}, middle). Next, we evaluate the arbitrage opportunities enabled by each distilled model when paired with Qwen Coder 72B. We measure mean profitability between 70\% and 90\% solve rate and plot it against the number of distillation tokens in Figure~\ref{fig:scaling} (right). We observe that increased distillation consistently creates more profitable arbitrage opportunities, with profitability increasing roughly log-linearly with the number of distillation tokens. The model distilled on 5B tokens (400k examples) enables a remarkably high level of profitability, allowing for a profit margin of nearly 30\%.

\end{document}